\documentclass[conference]{IEEEtran}
\IEEEoverridecommandlockouts
\usepackage{cite}
\usepackage{amsmath,amssymb,amsfonts}
\usepackage{algorithmic}
\usepackage{graphicx}
\usepackage{textcomp}
\usepackage{xcolor}
\usepackage[T1]{fontenc}
\usepackage[utf8]{inputenc}
\usepackage{moreverb,url}
\usepackage{graphicx}
\usepackage{subcaption} 
\usepackage{breakurl}
\usepackage[colorlinks,bookmarksopen,bookmarksnumbered,urlcolor=red]{hyperref}
\usepackage{csquotes}
\usepackage{textgreek}
\usepackage{tabularx}
\usepackage{xurl}

\newcommand\BibTeX{{\rmfamily B\kern-.05em \textsc{i\kern-.025em b}\kern-.08em
T\kern-.1667em\lower.7ex\hbox{E}\kern-.125emX}}

\usepackage{fancyhdr}

\fancypagestyle{IEEEcopyright}{
    \fancyhf{}
    \fancyfoot[C]{%
        \scriptsize
        \copyright~2026 IEEE. Personal use of this material is permitted. Permission from IEEE must be obtained for all other uses, in any current or future media, including reprinting/republishing this material for advertising or promotional purposes, creating new collective works, for resale or redistribution to servers or lists, or reuse of any copyrighted component of this work in other works.
    }

}

\begin{document}

\title{A Repeated-Measurement Study\\for Cultural Analytics of English Song Lyrics \\Using Five Large Language Models
\\}


\author{
\IEEEauthorblockN{
E. Cho Smith \hspace{3em}
Samuel Ho \hspace{3em}
Dawn Laux
}
\IEEEauthorblockA{
School of Applied and Creative Computing, Purdue University\\
West Lafayette, Indiana 47906-3603, USA\\
0009-0008-7817-690X \hspace{4em}
0000-0003-4207-2187 \hspace{4em}
0000-0001-6239-8307
}
}


\IEEEpubid{\makebox[\columnwidth]{978-1-6654-XXXX-X/26/\$31.00~\copyright~2026 IEEE\hfill}%
\hspace{\columnsep}\makebox[\columnwidth]{ }}

\maketitle

\thispagestyle{IEEEcopyright}

\begin{abstract}
Large language models (LLMs) are increasingly used to annotate cultural texts at scales that are impractical for human coders. However, before their outputs are treated as measurements of latent social constructs, it is necessary to establish whether those measurements are reliable. This study evaluates five LLMs as zero-shot annotators of four social constructs expressed in English song lyrics: self-esteem, self-control, seeking belonging, and seeking recognition. Using repeated annotations of a large lyric corpus, we examine three properties of LLM-based measurement: consistency across repeated runs, convergence across models, and transferability of consensus labels to supervised classification. The findings show that LLM-based measurement is not uniformly reliable across constructs. Self-esteem exhibits the strongest repeated-measurement reliability across models, while seeking recognition is generally less stable; self-control and seeking belonging show intermediate but model-dependent reliability. Downstream classification further indicates that consensus LLM labels contain learnable signal, although transferability does not itself establish construct validity. Repeated-measurement stability and cross-model convergence should therefore be reported before LLM annotations are treated as scalable measurements in cultural analytics.

\end{abstract}


\section{Introduction}
Cultural Analytics (CA) refers to the computational investigation of extensive cultural data to uncover patterns, diversity, and variation within cultural expression \cite{manovich2017cultural}. LLMs extend this computational capability by interpreting and categorizing meaning in large text corpora without requiring task-specific supervised training. However, scale itself does not establish measurement quality. When an LLM infers an unobserved construct from text, its output reflects a measurement produced by a particular model, model version, prompt, and annotation run. This issue becomes especially important when the target is not a directly observable lexical feature, but a latent social construct that must be inferred from meaning and context.

Recent studies have moved beyond treating song lyrics as purely aesthetic objects, examining them as signals of emotional processes and self-regulation strategies. Work on “coping songs” shows that lyrical themes can aid listener mood management and stress recovery \cite{levy2024lyrics}. Other work links lyrical themes with psychological risk factors such as depression \cite{chowdary2024lyrically}, while broader studies connect music preferences with emotional self-regulation \cite{de2023music}. Together, these findings highlight growing interest in modeling lyrics computationally, with LLMs offering new opportunities to analyze how lyrical signals map onto listeners’ self-regulatory needs. 

The present study evaluates four previously proposed dimensions for song-lyric annotation of latent social constructs: self-esteem, self-control, seeking belonging, and seeking recognition using frontier LLMs at the time \cite{fedcsissmith2024psychological, smith2024psychological}. This paper does not attempt to establish whether these four dimensions constitute the correct latent structure of lyrical meaning. That question concerns construct validity and is intentionally reserved for subsequent work after securing the reliability of uprising use of zero-shot learning using LLMs. Instead, three properties are examined. First, stability across repeated runs, convergence across LLM instruments, and downstream transferability of sufficiently stable labels to supervised classification. Together, these properties capture repeated-measurement reliability, cross-model convergence, and downstream transferability.

\section{Literature Review}
\subsection{LLMs as Measurement Instruments in Cultural Analytics}
CA seeks patterns across large-scale cultural data  \cite{manovich2017cultural}. When LLMs are used to assign semantic labels to cultural texts, the labels are not direct observations of the target construct; they are outputs of a generative instrument. The model version, prompt, decoding behavior, and annotation occasion are therefore part of the measurement specification. This framing motivates evaluating repeated measurements and cross-instrument convergence before treating LLM outputs as corpus-level evidence.

\subsection{Theoretical Rationale for the PNACSS Measurement Space}
The Psychological Needs as Credible Song Signals (PNACSS) framework was developed to operationalize recurring social-need signals in lyrics \cite{fedcsissmith2024psychological, smith2024psychological}. Its rationale draws primarily from two theoretical lines. First, the Temporal Need-Threat Model of ostracism proposes that social exclusion threatens fundamental needs including belonging, self-esteem, and control \cite{williams2009}. PNACSS adopts these needs as interpretable dimensions that may be expressed, threatened, restored, or negotiated in lyrical narratives. Second, work on music as credible and social signaling provides a rationale for treating songs as cultural artifacts through which social positioning and affiliation can be expressed \cite{mehr2021origins}. Research on music as a social surrogate further supports the idea that musical engagement can symbolically support social connection when direct interaction is limited \cite{schafer2020}.

PNACSS distinguishes self-esteem and self-control from the socially oriented constructs of seeking belonging and seeking recognition. Seeking belonging captures an expressed orientation toward connection, acceptance, or affiliation. Seeking recognition captures an expressed orientation toward acknowledgment, visibility, status, or validation. This is an operational coding framework rather than a claim that these dimensions exhaust the psychology of music listening. Music-in-everyday-life scholarship further supports treating musical meaning as socially situated rather than reducible to emotion labels alone \cite{denora2016music}.

\subsection{Why PNACSS Is an Appropriate Reliability Test Case}
The PNACSS dimensions require semantic interpretation beyond surface sentiment. Each lyric is mapped to four binary decisions whose combination yields 16 joint codes. If LLMs are to be used for CA of latent constructs, the same lyric should not receive substantially different measurements simply because the same model is queried again, and independent model instruments should show meaningful convergence when intended to measure the same construct. Reliability is not construct validity. A dimension can be measured consistently and still be theoretically incomplete or poorly aligned with the underlying human phenomenon. Accordingly, PNACSS is treated here as a fixed measurement target. Whether the dimensions themselves are valid latent constructs is reserved for subsequent validation research.

\section{The present study}
The present study examines whether LLMs can reliably annotate of latent social constructs in cultural texts, specifically English song lyrics, in which LLMs are not explicitly trained for this task. Previous studies explored this gap on a small scale, limited to about 5 to 12 lyrics across 16 categories \cite{fedcsissmith2024psychological, smith2024psychological}. Reported accuracy ranged from 53\% across all four dimensions to 93\% for a single dimension, leaving generalizability uncertain \cite{fedcsissmith2024psychological}. 

Thus, this study evaluates the reliability of LLMs in annotating latent psychological constructs within English song lyrics for four dimensions derived from theories of symbolic interaction and the need for regulation of self-esteem, self-control, seeking belonging, and seeking recognition. The current study addresses three research questions:

\textbf{RQ1: }To what extent are repeated LLM annotations reliable as measurements of the four target constructs in song lyrics, and how does within-model reliability vary by construct?

\textbf{RQ2: }To what extent do independently derived LLM consensus measurements converge on the same lyrics, and how does cross-model agreement vary by construct?

\textbf{RQ3: }Do consensus silver labels derived from a comparatively stable LLM measurement process contain transferable signal for supervised detection of the four constructs, including evaluation against human-validated annotations?

\section{Methods}
\subsection{Corpus and Preprocessing}
The annotation corpus was derived from Music4All database, a large-scale multi-modal corpus containing lyrics and associated musical metadata \cite{santana2020music4all}. From 109,269 entries, we filtered for English-language lyrics by applying two filters: (1) a length constraint retaining texts between 10 and 800 tokens. Token length was measured using the GPT-4o tiktoken encoding and (2) sentence-level language detection with Lingua, which excluded items where English was not the dominant language. After preprocessing, 69,130 lyrics (63.3\% of the original corpus) remained and formed the intended corpus for three annotation rounds across the five LLMs.

\subsection{LLMs and Annotation Protocol}
The 2025 experiment used five frontier LLMs: GPT-4o-mini (gpt-4o-mini-2024-07-18), o3-mini (o3-mini-2025-01-31), Claude 3.7 Sonnet (claude-3-7-sonnet-20250219), DeepSeek-R1, and Gemini 2.0 Flash (gemini-2.0-flash-001). Each model was run for three repeated rounds. The o1 model (o1-2024-12-17) was run only once and was therefore excluded from repeated-reliability analyses. All models were prompted under zero-shot conditions, meaning no examples were provided. All model queries were performed between March and May 2025; several of these model versions have since been retired or superseded. The reported agreement and transfer results should therefore be read as they were in 2025 and to the specific model versions tested, except for GPT-4o-mini, which is used to regenerate annotations in the third round. The original wrapper called the chat-completions API without fixed temperature, \texttt{top\_p, max\_tokens}, or seed. The original prompt was sent as a user message followed by the lyric as a second user message. 

\subsection{Psychological Needs as Credible Song Signals (PNACSS)}
\begin{figure}[htbp]
     \centering
     \includegraphics[width=0.50\linewidth]{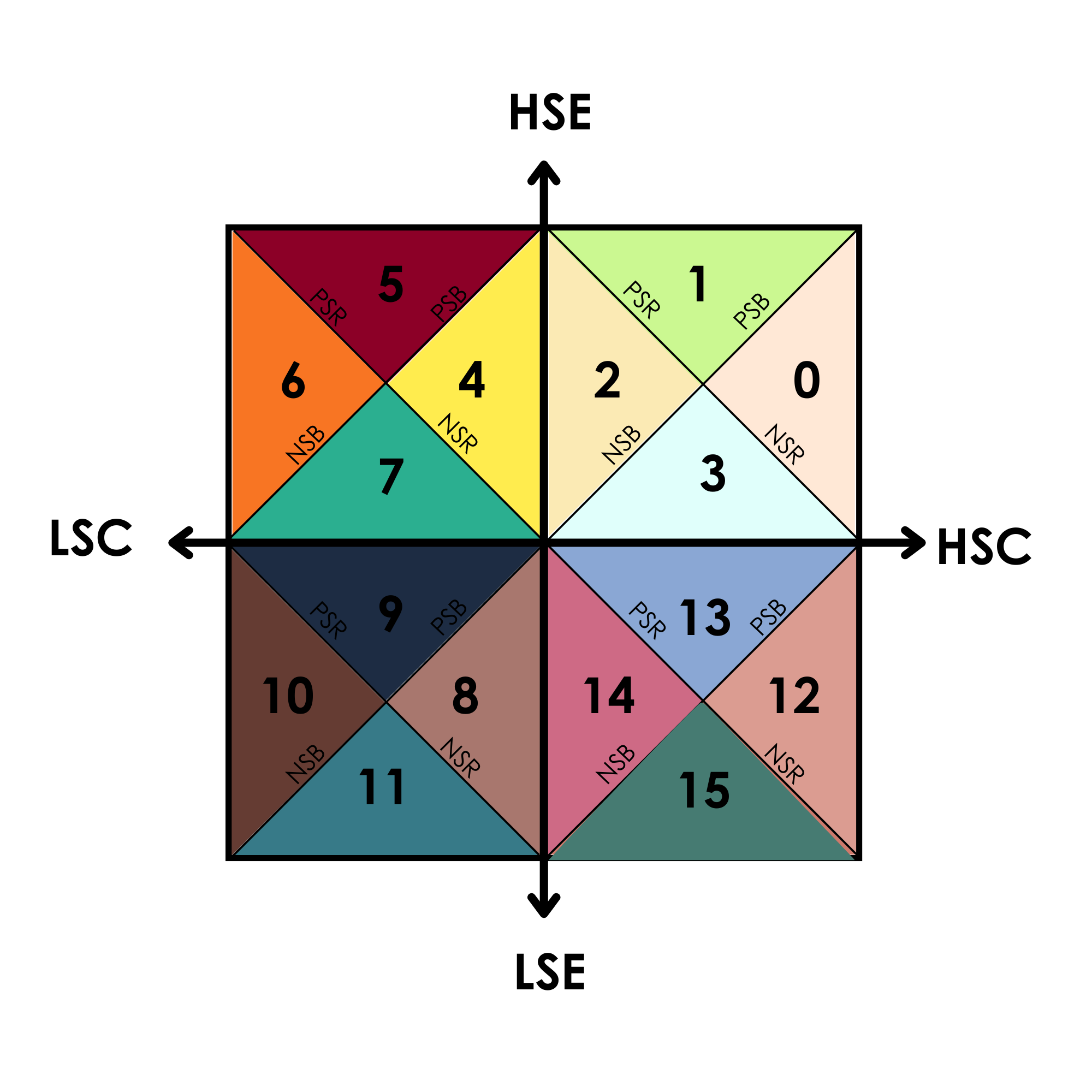}
     \caption{Psychological Needs as Credible Song Signals
  \cite{fedcsissmith2024psychological}.}
     \label{fig:pnmodel}
 \end{figure}

The annotation coding space was based on the PNACSS model shown in Fig.~\ref{fig:pnmodel}, an adaptation of the Temporal Need-Threat Model of Ostracism \cite{williams2009} to the domain of popular music. The PNACSS model identifies four constructs frequently negotiated in solitary listening: self-esteem, self-control, seeking belonging, and seeking recognition. 

\subsection{PNACSS Coding Space}
To code these needs, the model used two binary schemes: self-esteem (SE) and self-control (SC) were coded as high (H) or low (L), while seeking belonging (SB) and seeking recognition (SR) were coded as positive (P) or negative (N). As defined in \cite{fedcsissmith2024psychological}, these four binary dimensions produce 16 possible lyric categories. Table~\ref{tab:binary_coding} summarizes the coding scheme. Annotations were represented internally as four-character +/- strings ordered by self-esteem (SE), self-control (SC), seeking belonging (SB), and seeking recognition (SR).

\begin{table}[!t]
\centering
\caption{Binary Coding Scheme for Latent Constructs}
\label{tab:binary_coding}

\footnotesize
\begin{tabular*}{\columnwidth}
{@{\extracolsep{\fill}}lcc@{}}
\hline
\textbf{Dimension} &
\textbf{Binary States} &
\textbf{Code Position} \\
\hline
Self-Esteem         & HSE (+) / LSE (-) & 1 \\
Self-Control        & HSC (+) / LSC (-) & 2 \\
Seeking Belonging   & PSB (+) / NSB (-) & 3 \\
Seeking Recognition & PSR (+) / NSR (-) & 4 \\
\hline
\end{tabular*}
\end{table}

\subsection{Zero-shot annotation}
Annotations followed the PNACSS framework, which defines four binary PNACSS dimensions and 16 resulting categories \cite{fedcsissmith2024psychological}. The zero-shot prompt was identical to the annotation instruction used in prior work \cite{fedcsissmith2024psychological}:

\begin{quote}
\small
As a lyrics annotator, your task is to categorize the subjects' (the narrators or protagonists) states expressed or closely assumed in lyrics into one of the sixteen combinations based on four binary sub-dimensions and they are:
\begin{enumerate}
    \item The subjects can either have High Self-Esteem (HSE) or Low Self-Esteem (LSE).
    \item The subjects can either have High Self-Control (HSC) or Low Self-Control (LSC).
    \item The subjects can either have Seeking to Belong (PSB) or Not Seeking to Belong (NSB).
    \item The subjects can either have Seeking Recognition (PSR) or Not Seeking Recognition (NSR).
\end{enumerate} 

Therefore, when you classify each lyrics, you must determine the either-or sub-dimensions in four dimensions: self-esteem, self-control, seeking to belong, and seeking recognition.
\end{quote}

\subsection{GPT-4o-mini Round 3 Re-annotation}
The August 2026 audit found that the original GPT-4o-mini Round 3
directory retained 15,592 of 69,130 response files, leaving 53,538
items unavailable. Because gpt-4o-mini-2024-07-18 remained available, the missing items were re-annotated using the original prompt, message structure, and output conversion function. The Batch API was used for scale. For analysis, the surviving original Round 3 result was used whenever present; the re-annotated result was used only when the original was missing. Invalid outputs, including neutral labels, multiple classifications, refusals, or outputs that could not be converted unambiguously to one four-bit code, were excluded without retry. The completed GPT-4o-mini consensus pool contained 60,066 annotations.

Accordingly, the completed Round 3 dataset represents a combination
of original 2025 observations and 2026 re-annotated observations.
The re-annotated measurements should therefore be interpreted as a
provenance-controlled replication of the missing annotation process,
rather than recovery of the original 2025 outputs.

\subsection{Reliability and Consensus Rules}
For RQ1, Fleiss' $\kappa$ was computed on lyrics with three valid repeated measurements, with construct-specific reliability obtained by decomposing the four-character code into its four binary positions. For inter-model agreement, Fleiss' $\kappa$ was calculated separately by annotation round across the five LLMs. Consensus labels were retained when at least two valid repeated measurements agreed on the same four-bit label. This rule also permits a two-label consensus when exactly two valid measurements are available and agree. For RQ2, Cohen's $\kappa$ was used for the two-model Gemini--GPT-4o-mini comparison. The historical complete Gemini Round 3 item-level files were lost. The surviving data permitted recovery of 42,873 Gemini consensus labels. Matching these by lyric ID to the completed GPT-4o-mini consensus pool yielded 38,394 common lyrics.

\subsection{Downstream BigBird Transfer Experiment}
RQ3 reuses the historical BigBird experiments as a transferability test. BigBird-RoBERTa-base~\cite{zaheer2020big} was used. The joint model produced four sigmoid outputs with a .5 threshold, a learning rate of .00004, and three training epochs. As shown in Table~\ref{tab:internal-testsets-wide}, the balanced joint dataset contained 960 examples, with 60 samples per 16-category class, split into 768 training and 192 test examples.

Four separate binary classifiers were also trained, one for each social construct, using the same learning rate and three training epochs. Each binary dataset contained 1,000 examples, split into 800 training and 200 test examples (Table~\ref{tab:internal-testsets-wide}). Performance was then evaluated on the held-out external validation set summarized in Table~\ref{tab:external-prevalence-wide}.

\subsubsection{Internal Evaluation Sets}
The two internal evaluation protocols used can be summarized in Table \ref{tab:internal-testsets-wide}. 

\begin{table}[t]
\centering
\caption{Internal Evaluation Sets}
\label{tab:internal-testsets-wide}
\scriptsize
\begin{tabular*}{\columnwidth}
{@{\extracolsep{\fill}}lccc@{}}
\hline
\textbf{Dataset} & \textbf{n} & \textbf{Train/Test} \\
\hline
16-category balanced (joint 4-bit model) & 960 & 768/192 \\
Per-construct binary (four separate models) & 1,000 & 800/200 \\
\hline
\end{tabular*}

\vspace{0.3em}
\begin{minipage}{\columnwidth}
\scriptsize
\textit{Note.} The 16-category set contains 60 examples per joint
4-bit class. The binary protocol was applied separately to each of
the four constructs.
\end{minipage}
\end{table}

\subsubsection{External Validation Set}
The held-out external set contained 1,572 lyrics and preserved natural class imbalance (Table \ref{tab:external-prevalence-wide}). It was used after model selection to assess construct-level performance and detect over- or under-prediction.

\begin{table}[t]
\centering
\caption{External Validation Set Prevalence ($n=1{,}572$)}
\label{tab:external-prevalence-wide}
\scriptsize
\begin{tabular*}{\columnwidth}
{@{\extracolsep{\fill}}lcc@{}}
\hline
\textbf{Dimension} &
\textbf{Positive} &
\textbf{Negative} \\
\hline
Seeking Belonging
& 995 (63.3\%) & 577 (36.7\%) \\

Self-Control
& 654 (41.6\%) & 918 (58.4\%) \\

Self-Esteem
& 501 (31.9\%) & 1,071 (68.1\%) \\

Seeking Recognition
& 220 (14.0\%) & 1,352 (86.0\%) \\
\hline
\end{tabular*}
\end{table}

The external validation set consisted of human-validated annotations
assembled in prior work from curated and sampled lyric subsets
\cite{fedcsissmith2024psychological,smith2024psychological}. It was
reserved for final evaluation and therefore was not used for model
selection.

\subsection{Data availability \& Transparency}
The dataset analyzed in this study is Music4All\cite{santana2020music4all}, which contains copyrighted song lyrics. In compliance with copyright and licensing restrictions, neither the full corpus nor the derived annotations and code that directly process lyric text can be made publicly available. Researchers seeking access to the dataset should contact the original creators of Music4All to obtain it under the dataset's established licensing terms for research purposes.

\section{Findings}

\subsection{RQ1: Repeated-Measurement Reliability}

\begin{figure*}[ht]
    \centering
    \begin{subfigure}{0.32\textwidth}
        \centering
        \includegraphics[width=\linewidth]{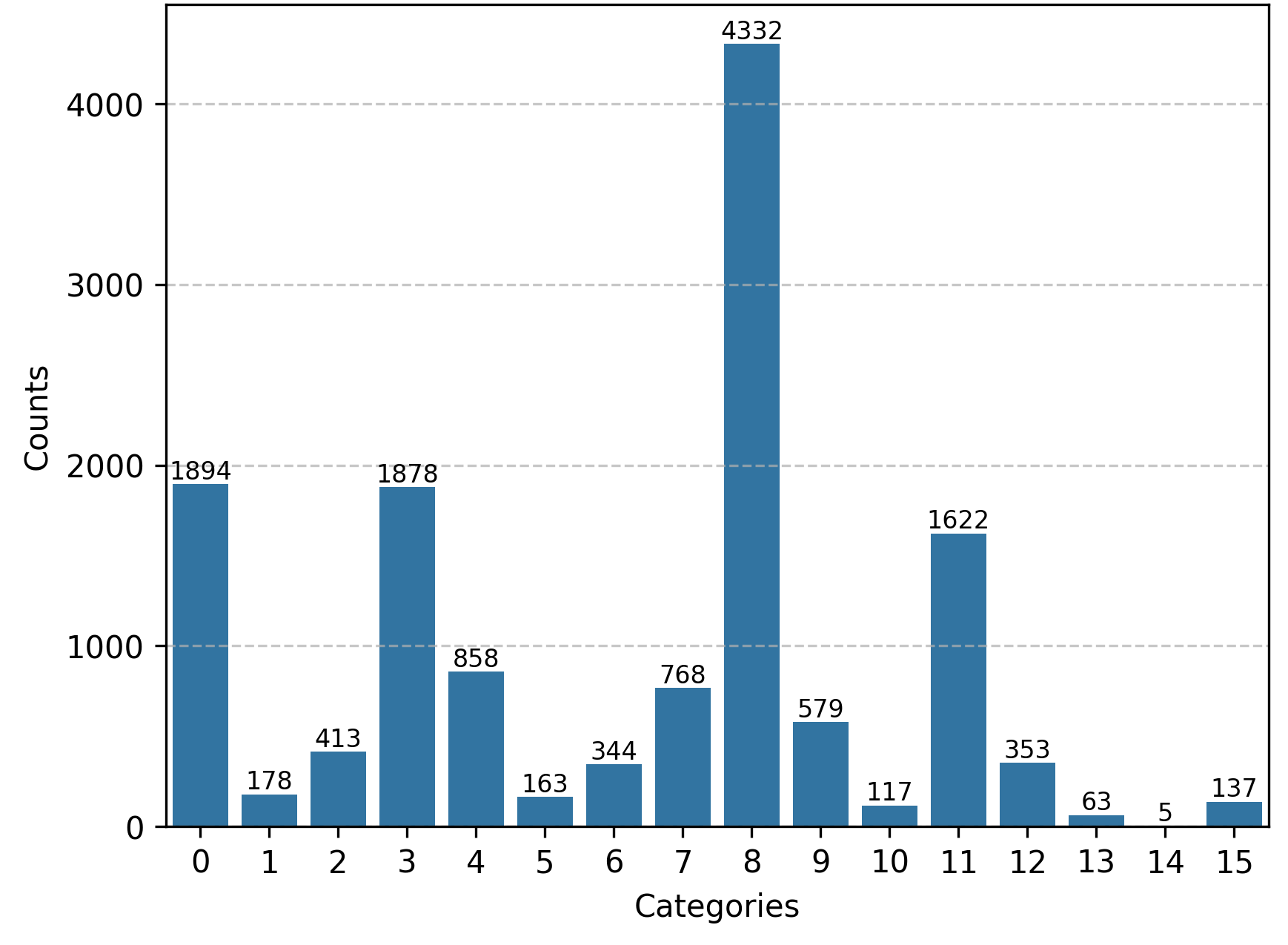}
        \caption{Model 1: o3-mini}
        \label{fig:model1_counts}
    \end{subfigure}
    \begin{subfigure}{0.32\textwidth}
        \centering
        \includegraphics[width=\linewidth]{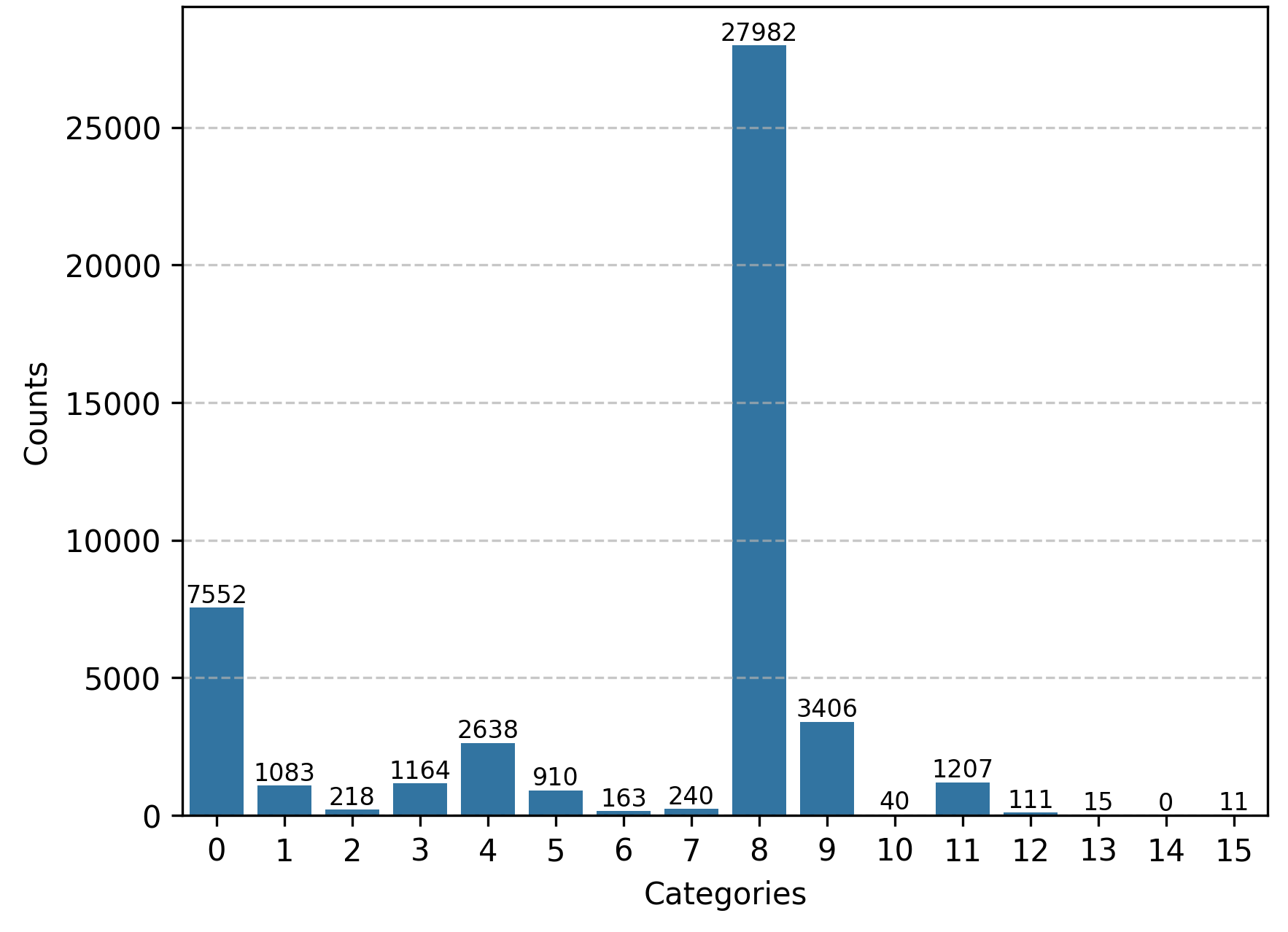}
        \caption{Model 2: GPT-4o-mini}
        \label{fig:model2_counts}
    \end{subfigure}
    \begin{subfigure}{0.32\textwidth}
        \centering
        \includegraphics[width=\linewidth]{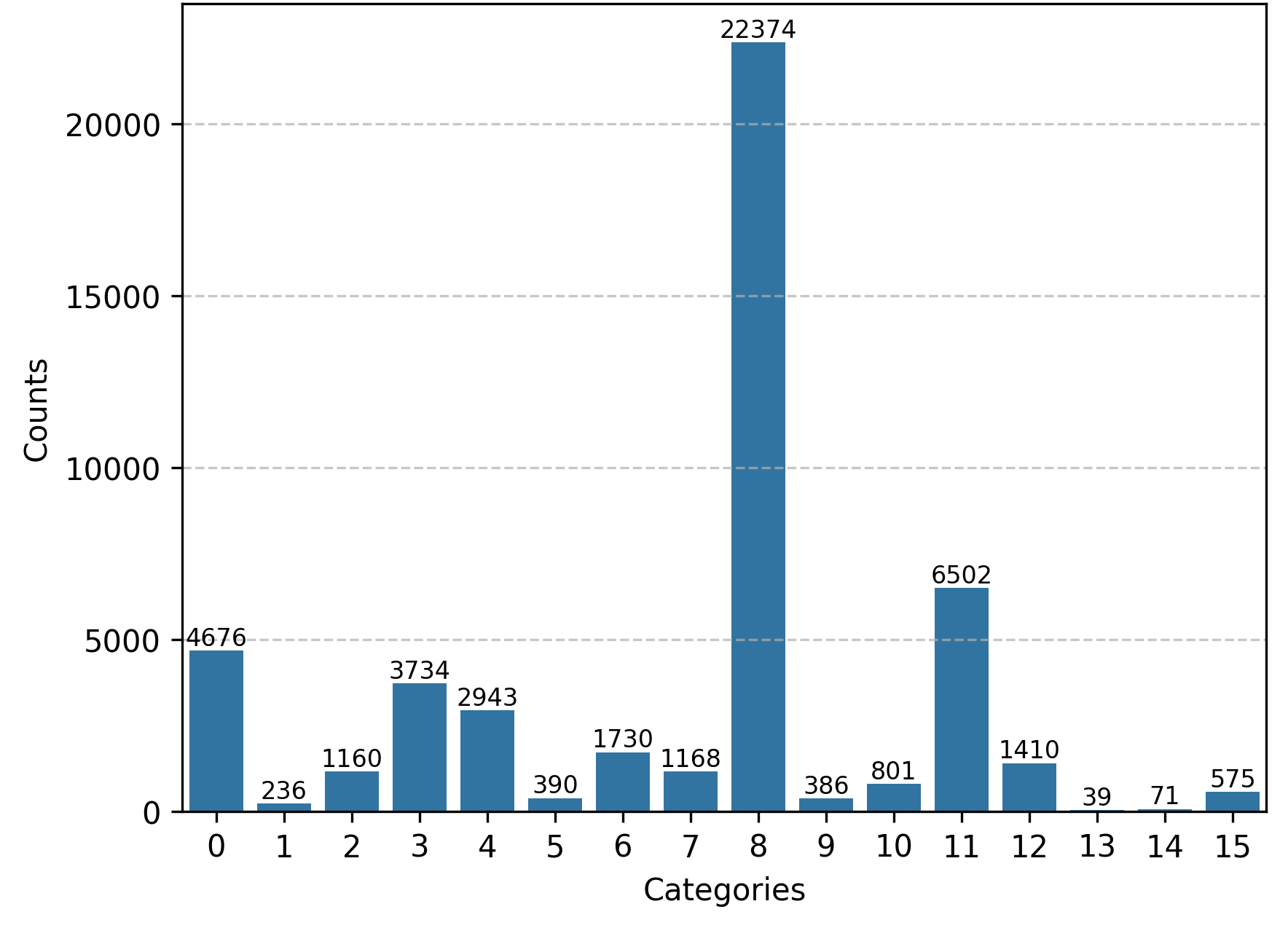}
        \caption{Model 3: DeepSeek-R1}
        \label{fig:model3_counts}
    \end{subfigure}
    \begin{subfigure}{0.32\textwidth}
        \centering
        \includegraphics[width=\linewidth]{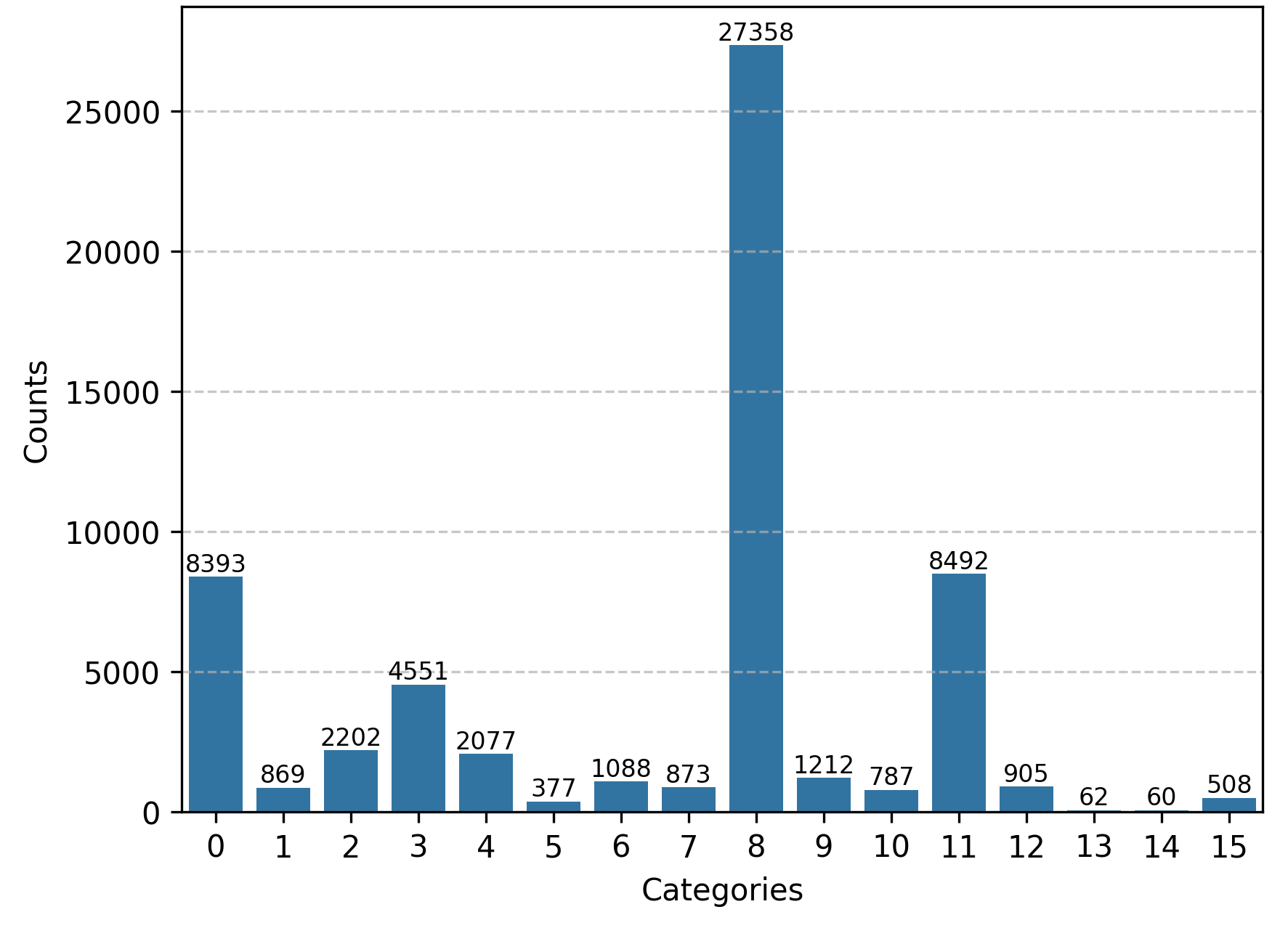}
        \caption{Model 4: Gemini 2.0 Flash}
        \label{fig:model4_counts}
    \end{subfigure}
    \begin{subfigure}{0.32\textwidth}
        \centering
        \includegraphics[width=\linewidth]{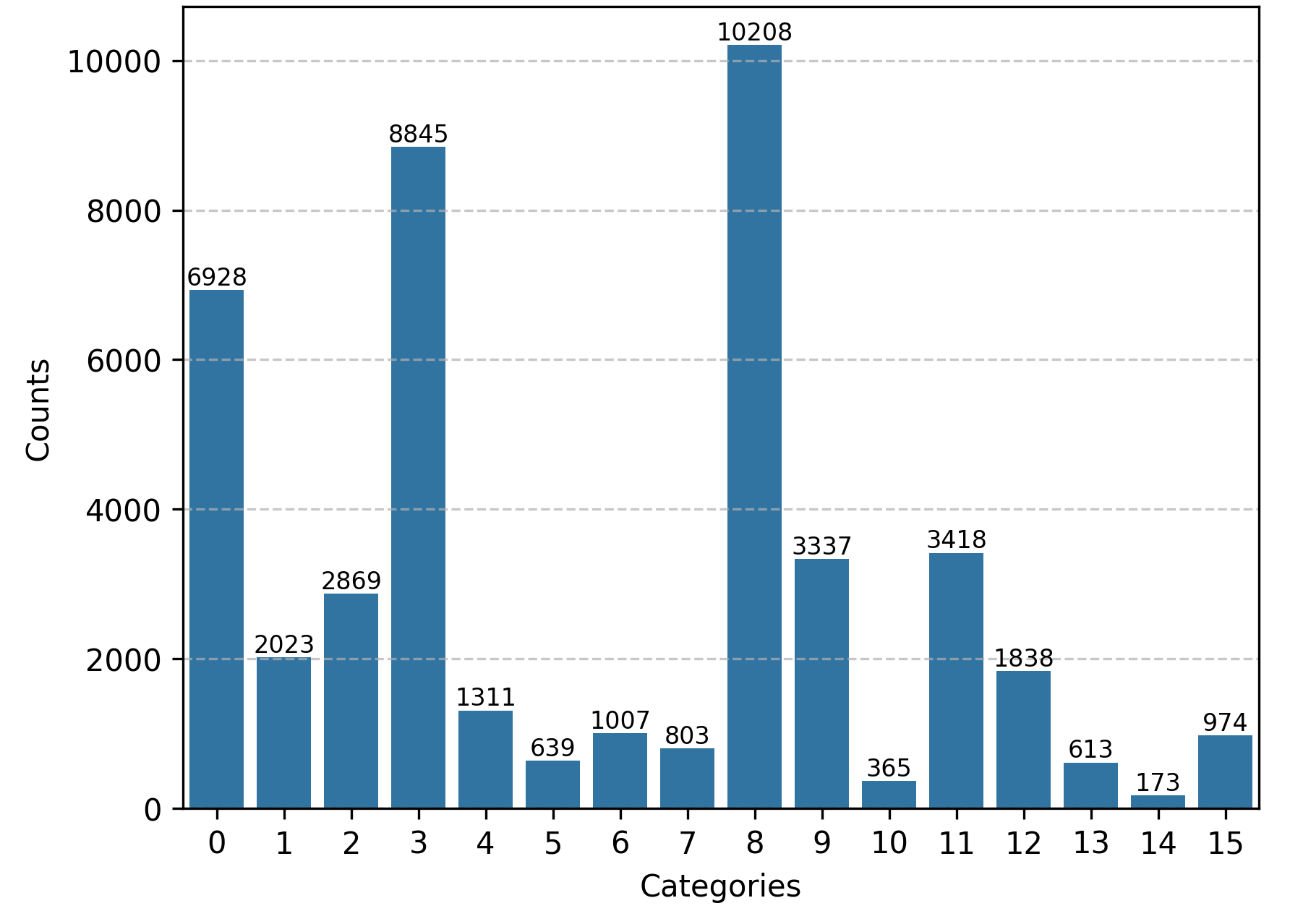}
        \caption{Model 5: Claude 3.7 Sonnet}
        \label{fig:model5_counts}
    \end{subfigure}
    \caption{Category distributions across the five models. Each chart shows counts for categories 0–15.}
    \label{fig:category_distributions}
\end{figure*}

Figure~\ref{fig:category_distributions} shows substantial differences in the models' consensus-label distributions. Category 8 (LLPN) was the modal category for all five models, although its prevalence varied considerably, indicating both shared and model-specific annotation bias.

Table~\ref{tab:intra_rater_by_model} shows that reliability was construct- and model-dependent. Self-esteem was consistently the most stable construct across all five models, with $\kappa$ values ranging from .717 to .826. Seeking recognition was generally the least stable, with values ranging from .467 to .683. Self-control and seeking belonging showed intermediate but more model-dependent stability, with their relative ordering varying across models. At the joint 16-category level, Fleiss' $\kappa$ ranged from .482 for Gemini 2.0 Flash to .614 for o3-mini. These results indicate that reliability cannot be adequately characterized by a single model-level or aggregate statistic; construct-level reliability is necessary when LLMs are used to measure individual dimensions.

\begin{table*}[!t]
\centering
\caption{Within-Model Reliability Within Each LLM Across Repeated Rounds}
\label{tab:intra_rater_by_model}
\scriptsize
\setlength{\tabcolsep}{3.5pt}
\renewcommand{\arraystretch}{0.92}

\begin{tabular*}{0.96\textwidth}
{@{\extracolsep{\fill}}llccccl@{}}
\hline
\textbf{LLM} &
\textbf{Construct / Classification} &
\textbf{n} &
\textbf{Fleiss' $\kappa$} &
\textbf{95\% CI} &
\textbf{SE} &
\textbf{Interpretation} \\
\hline

GPT-4o-mini
& \textbf{16-category}
& 55,982 & .517 & [.513, .521] & .002 & Moderate \\
& Self-Esteem
& 55,982 & .801 & [.796, .805] & .002 & Almost Perfect \\
& Self-Control
& 55,982 & .719 & [.714, .724] & .003 & Substantial \\
& Belonging
& 55,982 & .577 & [.568, .587] & .005 & Moderate \\
& Recognition
& 55,982 & .467 & [.459, .474] & .004 & Moderate \\
\hline

o3-mini
& \textbf{16-category}
& 4,583 & .614 & [.601, .626] & .006 & Substantial \\
& Self-Esteem
& 4,583 & .826 & [.813, .838] & .006 & Almost Perfect \\
& Self-Control
& 4,583 & .750 & [.733, .767] & .009 & Substantial \\
& Belonging
& 4,583 & .777 & [.762, .793] & .008 & Substantial \\
& Recognition
& 4,583 & .660 & [.633, .683] & .012 & Substantial \\
\hline

Claude 3.7 Sonnet
& \textbf{16-category}
& 1,454 & .584 & [.563, .604] & .011 & Moderate \\
& Self-Esteem
& 1,454 & .800 & [.776, .823] & .012 & Substantial \\
& Self-Control
& 1,454 & .763 & [.733, .790] & .014 & Substantial \\
& Belonging
& 1,454 & .741 & [.713, .769] & .015 & Substantial \\
& Recognition
& 1,454 & .683 & [.646, .717] & .018 & Substantial \\
\hline

DeepSeek-R1
& \textbf{16-category}
& 1,803 & .587 & [.565, .607] & .010 & Moderate \\
& Self-Esteem
& 1,803 & .810 & [.789, .832] & .012 & Almost Perfect \\
& Self-Control
& 1,803 & .687 & [.658, .714] & .015 & Substantial \\
& Belonging
& 1,803 & .734 & [.709, .757] & .013 & Substantial \\
& Recognition
& 1,803 & .651 & [.609, .694] & .022 & Substantial \\
\hline

Gemini 2.0 Flash
& \textbf{16-category}
& 9,094 & .482 & [.474, .492] & .005 & Moderate \\
& Self-Esteem
& 9,094 & .717 & [.704, .728] & .006 & Substantial \\
& Self-Control
& 9,094 & .651 & [.638, .666] & .007 & Substantial \\
& Belonging
& 9,094 & .600 & [.588, .615] & .007 & Moderate \\
& Recognition
& 9,094 & .604 & [.587, .624] & .010 & Moderate \\
\hline
\end{tabular*}

\vspace{0.25em}
\begin{minipage}{0.96\textwidth}
\scriptsize
\textit{Note.} Fleiss' $\kappa$ uses observations with three valid
repeated ratings. GPT-4o-mini includes the re-annotated Round 3.
Other reduced $n$ values reflect incomplete historical Round 3 data.
All 95\% CIs exclude zero. 
\end{minipage}
\end{table*}

As a two-round sensitivity check, Gemini showed the same construct-level
pattern (self-esteem $\kappa=.721$, self-control $.652$, belonging
$.593$, recognition $.600$), supporting the three-round results despite
its incomplete historical Round 3.

\subsection{RQ2: Cross-Model Agreement}

Cross-model agreement was consistently lower than within-model reliability (Table~\ref{tab:inter_rater}). Across all five models, self-esteem showed the strongest agreement in both rounds ($\kappa=.605$ and $\kappa=.608$), followed by self-control ($\kappa=.481$ and $\kappa=.479$), seeking belonging ($\kappa=.442$ in both rounds), and seeking recognition ($\kappa=.421$ and $\kappa=.419$). Thus, although all four constructs showed nontrivial convergence, the magnitude of agreement varied substantially by construct.

\begin{table}[!t]
\centering
\caption{Inter-Model Agreement Across Five LLMs}
\label{tab:inter_rater}
\scriptsize
\setlength{\tabcolsep}{2.5pt}

\begin{tabular*}{\columnwidth}
{@{\extracolsep{\fill}}clccl@{}}
\hline
\textbf{Rnd.} &
\textbf{Construct} &
\textbf{$\kappa$} &
\textbf{95\% CI} &
\textbf{Interpretation} \\
\hline
1 & Self-Esteem          & .605 & [.601, .609] & Substantial \\
1 & Self-Control         & .481 & [.477, .485] & Moderate \\
1 & Seeking Belonging    & .442 & [.438, .446] & Moderate \\
1 & Seeking Recognition  & .421 & [.416, .427] & Moderate \\
\hline
2 & Self-Esteem          & .608 & [.600, .615] & Substantial \\
2 & Self-Control         & .479 & [.471, .487] & Moderate \\
2 & Seeking Belonging    & .442 & [.434, .449] & Moderate \\
2 & Seeking Recognition  & .419 & [.409, .430] & Moderate \\
\hline
\end{tabular*}

\vspace{0.2em}
\begin{minipage}{\columnwidth}
\scriptsize
\textit{Note.} Fleiss' $\kappa$ compares the five LLMs.
Complete-case $n=68{,}583$ (Round 1) and $17{,}583$ (Round 2).
\end{minipage}
\end{table}

The direct comparison of Gemini and GPT-4o-mini consensus labels provided a complementary view. Agreement was strongest for self-esteem ($\kappa=.816$) and self-control ($\kappa=.752$), while seeking belonging showed substantially lower agreement ($\kappa=.337$). Seeking recognition fell between these extremes ($\kappa=.512$). Notably, raw agreement remained relatively high even when chance-corrected agreement was weaker, particularly for seeking belonging, illustrating why raw agreement alone is insufficient for evaluating convergence in imbalanced categorical annotations.

\begin{table}[!t]
\centering
\caption{Agreement Between Gemini and GPT-4o-mini}
\label{tab:cross_model_gemini_gpt}
\scriptsize
\setlength{\tabcolsep}{2.5pt}

\begin{tabular*}{\columnwidth}
{@{\extracolsep{\fill}}lcccl@{}}
\hline
\textbf{Measure} &
\textbf{$\kappa$} &
\textbf{Raw Agreement} &
\multicolumn{2}{c}{\textbf{Interpretation}} \\
\hline
Full 16-category
& .459 & 64.84\% & \multicolumn{2}{l}{Moderate} \\
Self-Esteem
& .816 & 92.33\% & \multicolumn{2}{l}{Almost Perfect} \\
Self-Control
& .752 & 90.96\% & \multicolumn{2}{l}{Substantial} \\
Seeking Belonging
& .337 & 80.33\% & \multicolumn{2}{l}{Fair} \\
Seeking Recognition
& .512 & 90.88\% & \multicolumn{2}{l}{Moderate} \\
\hline
\end{tabular*}

\vspace{0.2em}
\begin{minipage}{\columnwidth}
\scriptsize
\textit{Note.} $n=38{,}394$ common consensus-labeled lyrics.
Cohen's $\kappa$ is chance-corrected agreement.
\end{minipage}
\end{table}

\subsection{RQ3: Transferability of Silver Labels}

The joint BigBird model achieved macro-F1=.659, precision=.584, recall=.792, and exact 4-bit accuracy=.104, only modestly above the 6.25\% chance level for 16 balanced classes. The four independent binary classifiers showed stronger per-construct performance (Table~\ref{tab:binary_classifier_metrics}), with macro-F1 ranging from .684 to .732.

\begin{table}
\centering
\footnotesize
\caption{Balanced Binary Classifier Performance}
\label{tab:binary_classifier_metrics}
\begin{tabular*}{\textwidth}{@{\extracolsep{\fill}} l c c c c c}
\hline
\textbf{Construct} &
\textbf{Macro-F1} &
\textbf{Precision} &
\textbf{Recall} &
\textbf{Accuracy} &
\textbf{Test Loss}\\
\hline
Self-Esteem         & .732 & .771 & .740 & .740 & .598 \\
Self-Control        & .696 & .712 & .700 & .700 & .657 \\
Seeking Belonging   & .698 & .707 & .700 & .700 & .657 \\
Seeking Recognition & .684 & .687 & .685 & .685 & .580 \\
\hline
\end{tabular*}
\end{table}

Under natural class imbalance (Table~\ref{tab:external_validation}), the classifiers displayed substantially different precision--recall profiles. Seeking belonging retained high precision (.679) and very high recall (.969), whereas self-control (precision=.447, recall=.928) and especially seeking recognition (precision=.207, recall=.886) showed strong positive-class over-prediction. Self-esteem showed the most balanced behavior among the four constructs (precision=.524, recall=.559). These results indicate that the silver-label signal is learnable, but that transfer to naturally imbalanced data can amplify the prevalence and labeling biases present in the training annotations.

\begin{table}[!t]
\centering
\caption{External Validation Performance ($n=1{,}572$)}
\label{tab:external_validation}
\scriptsize

\begin{tabular*}{\columnwidth}
{@{\extracolsep{\fill}}lcc@{}}
\hline
\textbf{Construct} &
\textbf{Precision} &
\textbf{Recall} \\
\hline
Self-Esteem         & .524 & .559 \\
Self-Control        & .447 & .928 \\
Seeking Belonging   & .679 & .969 \\
Seeking Recognition & .207 & .886 \\
\hline
\end{tabular*}
\end{table}

\begin{figure*}[t]
\centering

\begin{subfigure}[t]{0.18\textwidth}
  \includegraphics[width=\linewidth]{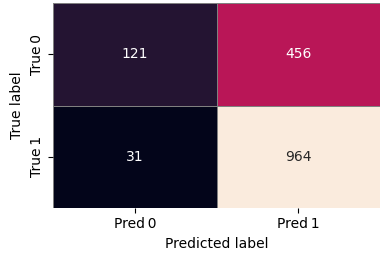}
  \caption{Seeking Belonging}
\end{subfigure}\hfill
\begin{subfigure}[t]{0.18\textwidth}
  \includegraphics[width=\linewidth]{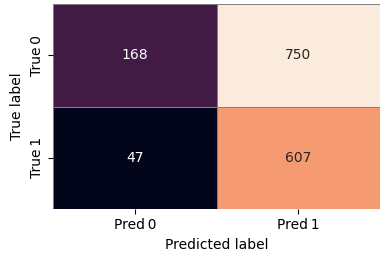}
  \caption{Self-Control}
\end{subfigure}\hfill
\begin{subfigure}[t]{0.18\textwidth}
  \includegraphics[width=\linewidth]{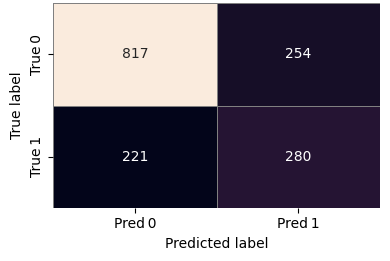}
  \caption{Self-Esteem}
\end{subfigure}\hfill
\begin{subfigure}[t]{0.18\textwidth}
  \includegraphics[width=\linewidth]{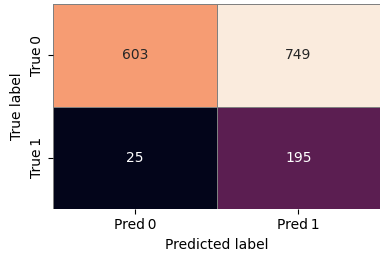}
  \caption{Seeking Recognition}
\end{subfigure}

\caption{Confusion matrices for the external validation set ($n=1{,}572$).}
\label{fig:external_confusion}
\end{figure*}

Figure~\ref{fig:external_confusion} shows the corresponding confusion matrices and confirms that high recall was often obtained by over-predicting positive cases.



\section{Discussion}

The principal finding is that LLM annotation reliability is construct-dependent. Self-esteem was consistently the most stable construct, seeking recognition was generally less stable, and self-control and seeking belonging showed more model-dependent stability. Cross-model agreement was consistently weaker than within-model reliability, indicating that reproducibility within a single LLM does not imply convergence across LLM instruments. These results suggest that cultural analytics should report reliability at the construct level rather than summarize annotation quality with a single aggregate statistic.

The strong concentration in Category 8 provides additional evidence of systematic annotation bias. Although a shared modal category does not establish invalidity, such concentration reduces the interpretive resolution of the silver labels. Downstream classifiers nevertheless learned meaningful signal, but their behavior under natural class imbalance showed that label learnability does not establish construct validity. In particular, very high recall and low precision for seeking recognition indicate positive-class over-prediction.

The GPT-4o-mini audit also highlights the importance of model and data provenance. More than 53,000 Round 3 responses were unavailable and had to be re-generated using the same model version and annotation procedure. These outputs are not recoveries of the original 2025 measurements, but provenance-controlled replications of the missing annotation process. Gemini's unrecoverable Round 3 similarly constrained cross-model analysis.

Three limitations are important. First, the original annotation wrapper did not fix temperature, top-$p$, or random seed, so repeated measurements reflect both model behavior and provider defaults. Second, the external validation set derives from prior human-in-the-loop annotation and is not independent behavioral or psychological ground truth. Third, this study evaluates reliability of the PNACSS coding space, not its construct validity. Future work should compare these dimensions with independent human judgments, behavioral measures, listening behavior, and alternative operationalization of psychological needs.

\section{Conclusion}

This study evaluated large-scale LLM annotation of song lyrics as a measurement problem in CA. Across repeated annotation rounds and multiple model systems, reliability varied substantially by construct. Self-esteem showed the strongest and most consistent reliability, seeking recognition was generally less stable, and self-control and seeking belonging exhibited model-dependent patterns. Cross-model agreement was consistently weaker than within-model stability, indicating that reproducibility within a single LLM should not be interpreted as evidence that different LLMs recover the same measurement.

The downstream experiments provide an important complementary result: LLM-generated silver labels contain learnable predictive signal, but their transferability does not establish construct validity. Strong recall combined with low precision under natural class imbalance shows that classifiers can reproduce systematic tendencies in the silver labels rather than recover an unbiased representation of the target constructs. LLM annotations should therefore be treated as weak supervision rather than ground truth unless supported by independent validation.

Accordingly, LLM-based cultural analytics should report model version, prompt, annotation configuration, repeated-measurement reliability, cross-model convergence, data provenance, and independent validation before treating generated labels as corpus-level evidence.

\section*{Acknowledgment}

In memory of Yinxuan ``Owen'' Wang, whose unwavering spirit and experimental work sustained this project. We dedicate this article to his radiant life.

\bibliographystyle{IEEEtran}
\bibliography{references}

\end{document}